\documentclass[10pt,twocolumn,letterpaper]{article}

\usepackage{cvpr}
\usepackage{times}
\usepackage{epsfig}
\usepackage{graphicx}
\usepackage{amsmath}
\usepackage{amssymb}
\usepackage{color}
\usepackage{verbatim}

\usepackage[breaklinks=true,bookmarks=false]{hyperref}

\cvprfinalcopy 

\def\cvprPaperID{****} 

\begin{document}

\title{Encoding Video and Label Priors for Multi-label Video Classification \\on YouTube-8M dataset}

\author{Seil Na\\
Seoul National University\\
{\tt\small seil.na@vision.snu.ac.kr}
\and
YoungJae Yu\\
Seoul National University\\
{\tt\small yj.yu@vision.snu.ac.kr}
\and
Sangho Lee\\
Seoul National University\\
{\tt\small sangho.lee@vision.snu.ac.kr}
\and
Jisung Kim\\
SK Telecom Video Tech. Lab\\
{\tt\small joyful.kim@sk.com}
\and
Gunhee Kim\\
Seoul National University\\
{\tt\small gunhee@snu.ac.kr}
}

\maketitle

\begin{abstract}
YouTube-8M is the largest video dataset for multi-label video classification.
In order to tackle the multi-label classification on this challenging dataset, it is necessary to solve several issues such as temporal modeling of videos, label imbalances, and correlations between labels.
We develop a deep neural network model, which consists of four components: Video Pooling Layer, Classification Layer, Label Processing Layer, and Loss Function.
We introduce our newly proposed methods and discusses how existing models operate in the YouTube-8M Classification Task, what insights they have, and why they succeed (or fail) to achieve good performance.
Most of the models we proposed are very high compared to the baseline models, and the ensemble of the models we used is 8th in the Kaggle Competition.
\end{abstract}

\section{Introduction}
\label{sec:intro}

\begin{figure}[t]
\centering
\includegraphics[trim=0.2cm 0.2cm 0cm 0.1cm,clip,width=0.50\textwidth]{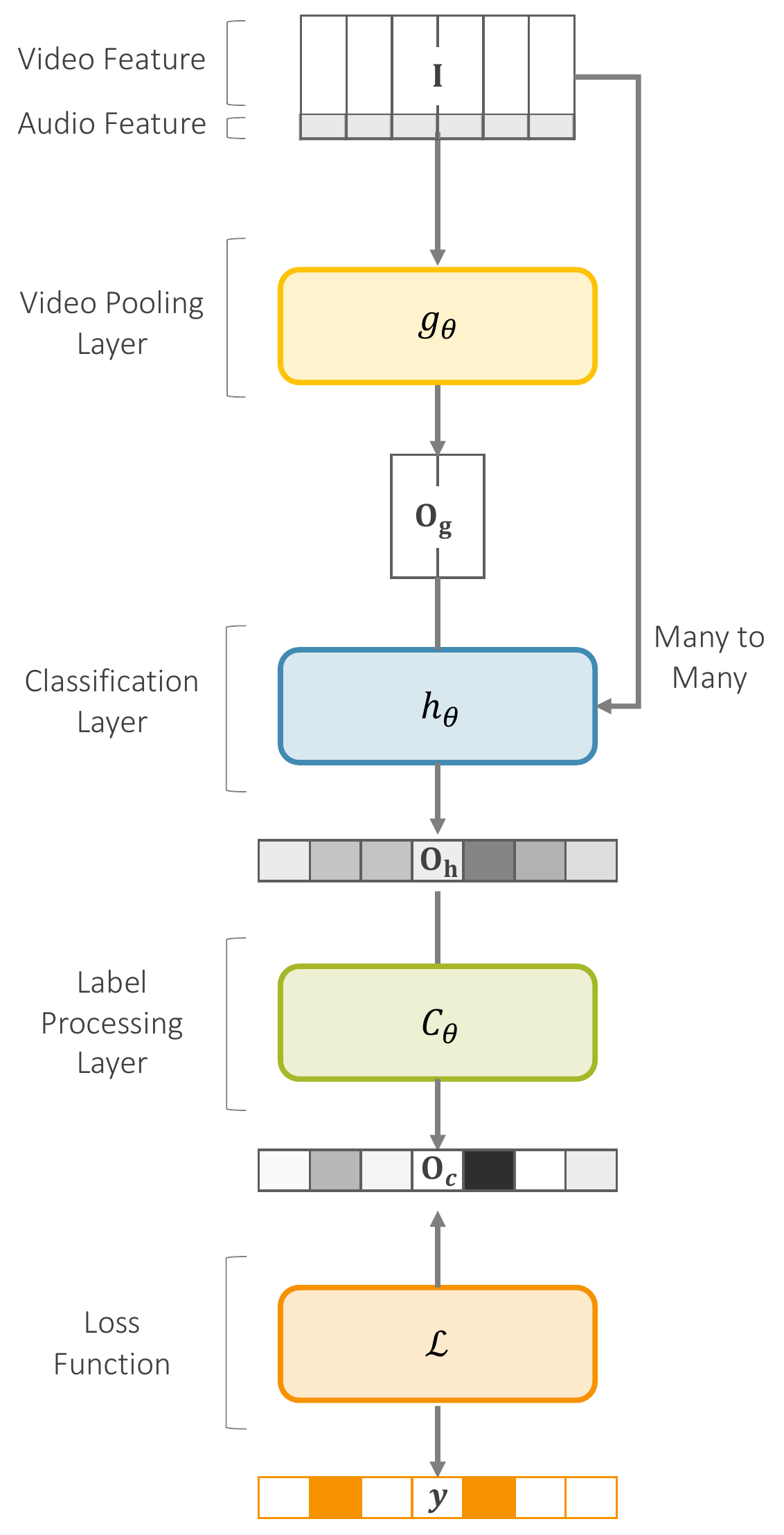}
\vspace{6pt}
\caption{A complete pipeline illustration of our model for YouTube-8M classification. The input of the model concatenates the frame feature $\mathbf{I^f}$ and audio feature $\mathbf{I^a}$, which is denoted by $\mathbf{I}$. We divided the model into four components (Video Pooling Layer, Classification Layer, Label Processing Layer, Loss Function) and experimented with the newly proposed model or variants of previously proposed model in each component. Each of the four components is denoted by $(g_\theta, h_\theta, C_\theta, \mathcal{L})$, and each output of first 3 layer is denoted by ($\mathbf{O_g}, \mathbf{O_h}, \mathbf{O_c}$).
}
\label{fig:overall}
\end{figure}

Many challenging problems have been studied in computer vision research toward video understanding, such as video classification~\cite{karpathy2014large,soomro2012ucf101}, video captioning~\cite{xu2016msr, yu2016video}, video QA~\cite{zhu2015uncovering}, and MovieQA~\cite{tapaswi2016movieqa}, to name a few.
YouTube-8M~\cite{abu2016youtube} is the largest video dataset for multi-label video classification. 
Its main problem is to predict the most relevant labels for a given video out of 4,716 predefined classes.
Therefore, it requires jointly solving two important problems; video classification and multi-label classification.

From the view of the video classification, YouTube-8M is challenging in that it covers more general classes like soccer, game, vehicle, and food, 
while existing video classification datasets focus on more specific class groups, such as sports in Sports-1M~\cite{karpathy2014large}, and actions in UCF-101~\cite{soomro2012ucf101}.
Therefore, unlike the importance of modeling motion features in UCF-101~\cite{soomro2012ucf101} or Sports-1M~\cite{karpathy2014large}, it is important to capture more generic video information(\eg temporal encoding method for video, audio feature modeling) in YouTube-8M.

From the view of multi-label classification, the key issues to solve in YouTube-8M are label imbalances and correlations between labels.
YouTube-8M involves 4,716 class labels, and the number of videos belonging to each class is significantly different, which causes a label imbalance issue that the classifier fits to the biased data.
At the same time, many classes are closely related one another, such as \{\textit{Football, Kick, Penalty kick, Indoor soccer}\} or \{\textit{Super Mario Bros, Super Mario World, Super Mario bros 3, Mario Kart, Mario Kart 8}\}. It is also challenging to resolve the correlations between labels to decide final prediction.

Based on the challenges of the multi-label video classification task on YouTube-8M described the above,
we focus on addressing i) temporal encoding for video, ii) relieving the label imbalance problem, and iii) utilizing the correlated label information.
Our model consists of four components: i) video pooling layer, ii) classification layer, iii) label processing layer, and iv) loss function.
The proposed components indeed show significant performance improvement over the baseline models of YouTube-8M~\cite{abu2016youtube}, 
and finally our ensemble model is ranked 8th in the Google Cloud \& YouTube-8M Video Understanding Challenge~\footnote{\url{https://www.kaggle.com/c/youtube8m/leaderboard}. (Team name: SNUVL X SKT)}.

\section{The Model}
\label{sec:model}

Figure \ref{fig:overall} shows the overall pipeline of our model. 
We first present video features we used, and then explain its four key components in the following sections. 

\subsection{Video Features}
\label{sec:feature_extraction}

The inputs of the model are frame features and audio features of a video clip.
The frame features are obtained by sampling a clip at 1-second interval, and extracting 2,048-dimensional vector from every frame through Inception-V3 Network~\cite{szegedy2016rethinking} pretrained to ImageNet~\cite{deng2009imagenet}.
Then the feature is reduced to an 1,024-dimension via PCA (+ whitening), then quantized, and finally L2-normalized.
As a result, for a video of $T$ seconds long, its frame is $\mathbf{I^f} \in \mathbb{R}^{T \times 1,024}$.
The audio features re extracted using the VGG-inspired acoustic model~\cite{hershey2016cnn} followed by L2-normalization, which is denoted by $\mathbf{I^a} \in \mathbb{R}^{T \times 128}$.

As the input of our model, we concatenate the frame feature $\mathbf{I^f}$ and the audio feature $\mathbf{I^a}$ at every time step, denoted by $\mathbf{I} \in \mathbb{R}^{T \times 1,152}$.
We test the compact bilinear pooling~\cite{fukui2016multimodal} with various dimensions between $\mathbf {I^f}$ and $\mathbf {I^a}$, but all of them have significantly lower performance than the simple concatenation.
From now on, we use $\mathbf{I}$ to denote the input features of a video over all frames, and $\mathbf{I}_t \in \mathbb{R}^{1,152}$ as $t$-th frame vector.


\subsection{Video Pooling Layer}
\label{sec:frame_encoder}

The Video Pooling Layer $g_\theta: \mathbb{R}^{T \times 1,152} \to \mathbb{R}^{d}$ is defined as a parametric function that encodes a sequence of $T$ feature vectors $\mathbf{I}$ into a $d$-dimensional embedding vector. We test five different encoding structures as follows.

\subsubsection{A Variant of LSTM}
\label{sec:lstm}

The Long Short Term Memory (LSTM) model~\cite{hochreiter1997long} is one of the most popular frameworks for modeling sequence data.
We use a variant of the LSTM as follows:
\begin{align}
\label{eq:lstm1}
\mathbf i_t &= \sigma(\mathbf{I}_t \mathbf U^i + \mathbf s_{t-1} \mathbf W^i + \mathbf b_i) \\
\label{eq:lstm2}
\mathbf f_t &= \sigma(\mathbf{I}_t U^f+ \mathbf s_{t-1} \mathbf W^f + \mathbf b_f) \\
\label{eq:lstm3}
\mathbf o_t &= \sigma(\mathbf{I}_t U^o+ \mathbf s_{t-1} \mathbf W^o + \mathbf b_o) \\
\label{eq:lstm4}
\mathbf g_t &= \mbox{tanh}(\mathbf{I}_t \mathbf U^g+ \mathbf s_{t-1} \mathbf W^g + \mathbf b_g) \\
\label{eq:lstm5}
\mathbf c_t &= \mathbf c_{t-1} \circ \mathbf f_t + \mathbf g_t \circ \mathbf i_t \\
\label{eq:lstm5}
\mathbf s_t &= \mbox{tanh}( \mathbf c_t) \circ \mathbf o_t
\end{align}
\noindent where $t$ denotes each time step, $\mathbf i, \mathbf f, \mathbf o$ are the input, forget, output gate, $\mathbf c_t, \mathbf s_t$ are long-term and short-term memory respectively.

The baseline model uses only the final hidden states of the LSTM (\ie $c_T, s_T$), but we additionally exploit the following two states in order to extract as much information as possible from the LSTM: i) the state $\mathbf{M}_l = \sum_{t=1}^{T} \mathbf{I}_t$: the summation of the input feature $\mathbf{I}_t$ of each time step, and ii) the state $\mathbf{O}_l = \sum_{t=1}^T \mathbf  g_t$: the summation of the output of each time step of the LSTM.
We concatenate $\mathbf{M}_l$ and $\mathbf{O}_l$ with $c_T$ and $s_T$.
That is, if the cell size of the LSTM is $d$, the output $\mathbf g_\theta$ of the baseline that uses $c_T, s_T$ becomes a $(2 \times d)$-dimensional vector,
whereas our model that additionally uses $\mathbf{M}_l$ and $\mathbf{O}_l$ has the output of a $(4 \times d)$-dimensional vector.
Experimentally, we choose the LSTM cell size $d$ = 1,152.
We also apply the layer normalization~\cite{ba2016layer} to each layer of the LSTM for fast convergence, and also use the dropout with dropout rate=0.8 to increase the generalization capacity.

\begin{figure}[t]
\centering
\includegraphics[trim=0.2cm 0.2cm 0cm 0.1cm,clip,width=0.5\textwidth]{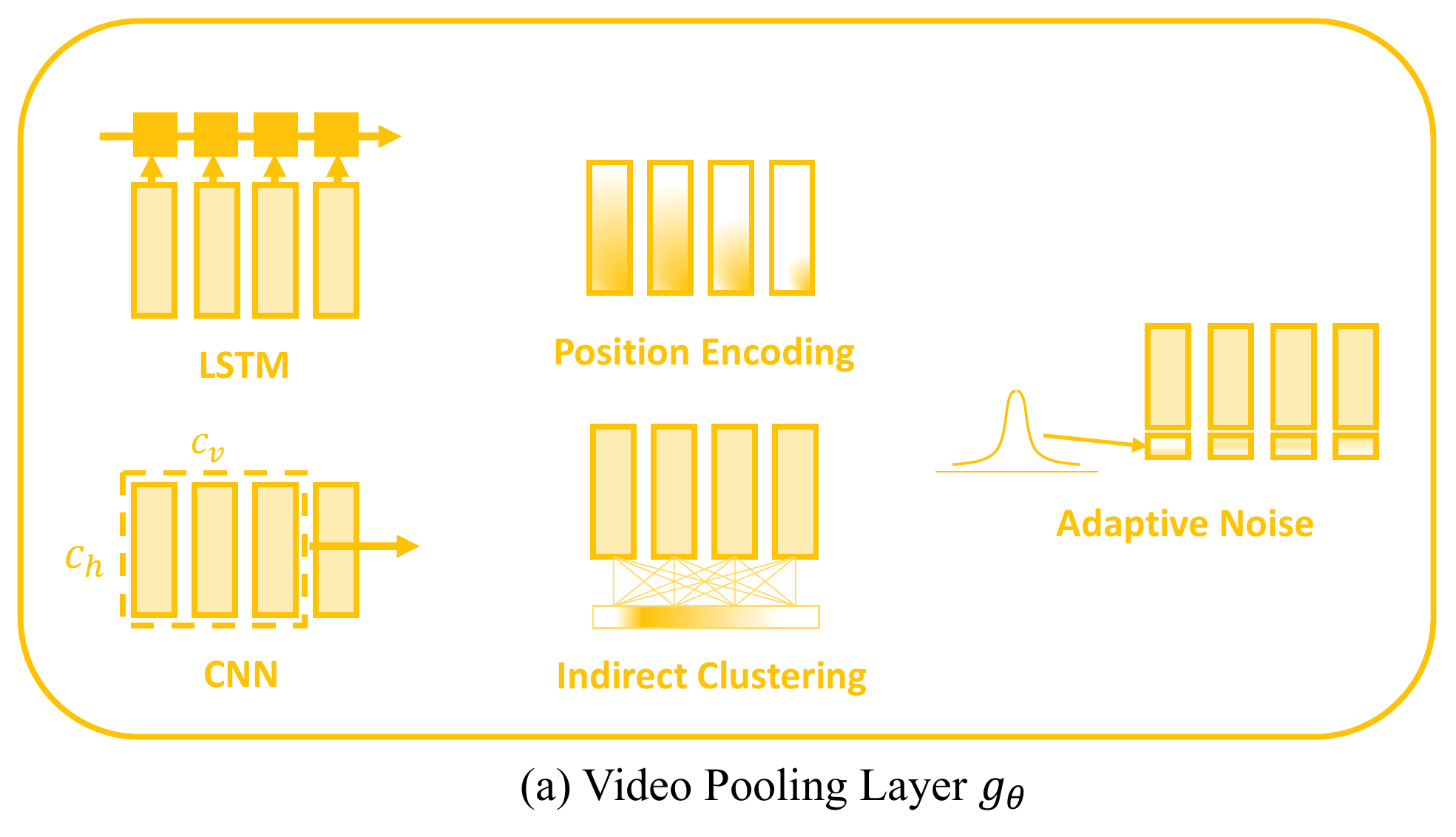}
\vspace{6pt}
\caption{The first component Video Pooling Layer $g_\theta$ of the entire model. It takes $T$ frame vector $\mathbf{I}$ as input and outputs $d$-dimensional vector using different encoding methods respectively.
}
\label{fig:frame_encoder}
\end{figure}
\begin{figure*}
\begin{center}
    \includegraphics[width=0.99\textwidth]{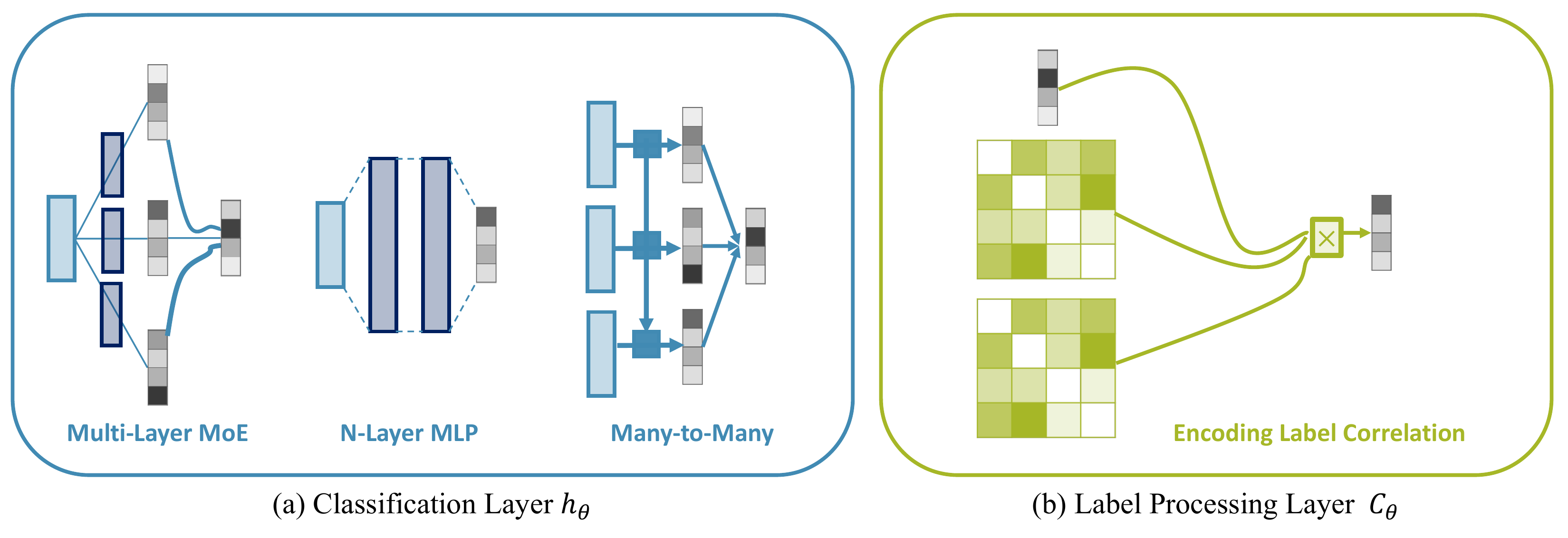}
\end{center}
\vspace{6pt}
   \caption{(a)The second component Classification Layer $h_\theta$ of the overall model. It take the $d$-dimensional vector as input $\mbox{O}_g$ and have score $\mbox{O}_h$ for 4,716 classes as output. However, input of Many-to-Many model is $\mbox{I}$ rather than $\mbox{O}_g$. (b)the 3rd component Label Processing Layer $C_\theta$ of the entire model. It uses label correlation prior to update the score for each class.
}
\label{fig:class_layer}
\end{figure*}

\subsubsection{CNNs}

Convolutional neural networks (CNNs) are often used to jointly capture spatial information from images or video in many computer vision tasks.
That is, the convolution kernels generate output signals considering all the elements in the window together, and thus they effectively work with spatially or temporally sequential information (\eg images, consecutive characters in NLP, and audio understanding).
As the second candidate of Video Pooling Layer, we use the CNN to capture temporal information of video as proposed in \cite{kim2014convolutional}:
\begin{equation}
\label{eq:cnn}
\mathbf{o}_c = \mbox{ReLU}(\mbox{conv}(\mathbf{I}, \mathbf{w}_{conv}, \mathbf{b}_{conv}))
\end{equation}
where conv(input, filter, bias) indicates convolution layer with stride 1, and ReLU indicates the element-wise ReLU activation~\cite{nair2010rectified}.
$\mathbf{w}_{conv} \in \mathbb{R}^{{c_v} \times {c_h} \times 1 \times d}$ is a convolution filter with the vertical and horizontal filter size of $c_v=5, c_h=1152$, and $\mathbf{b}_{conv} \in \mathbb{R}^d$ is a bias. $\mathbf{o}_c \in \mathbb{R}^{(T-c_v+1) \times d}$ indicates the output of the convolution layer.
Finally, we apply max-pooling over time for $\mathbf{o}_c$,  obtaining the $d$-dimensional encoding of $\mathbf g_\theta$.

\subsubsection{Position Encoding}
We also test the Position Encoding scheme~\cite{sukhbaatar2015end} that assigns different weights to each frame vector.
That is, we define the matrix $\mbox{L} \in \mathbb{R}^{T \times 1152}$, Position Encoding is simply defined as follows.
\begin{align}
\label{eq:position_encoding}
\mathbf{I} \leftarrow \mathbf{I} \odot \mbox{L} \\
\label{eq:position_encoding2}
O_g = \sum_{t=1}^{T} {\mathbf{I}_t}
\end{align}
where $\mbox{L}_{ij} = (1 - i/T) - (j/1152)(1 - 2 \times i/T)$, $\odot$ means element-wise multiplication.  
After applying Position Encoding, we used summation of each vector as output of $g_\theta$, that menas $O_g$ is $d$-dimensional vector where d is 1,152.

\subsubsection{Self-Attention: Indirect Clustering}
The YouTube-8M dataset deals with general topics (\eg soccer, game, car, animation) rather than relatively focused labels like Sports-1M or UCF-101.
We here test the following hypothesis: since the topic is highly general, it may be more advantageous to focus on the most dominant parts of the video rather than temporal/motion information of individual frames. 
Therefore, we suggest an indirect clustering model using the self-attention mechanism as follows. 

We perform a clustering on the video features over all frames $\mathbf{I} \in \mathbb{R}^{T \times 1,152}$, 
and find the cluster with the largest size (\ie the largest number of elements in the cluster).
Then, the vectors in this cluster may represent the main scene of the video.
However, since it takes very long time to perform clustering on each video, we propose a self-attention mechanism that acts like clustering as follows.
\begin{equation}
\label{eq:self_attention}
\mbox{p}_t = \mbox{softmax}(\sum_{i=1}^T \mbox{I}_t \mbox{I}_i)
\end{equation}
where $\mbox{p}_t$ is a scalar value that indicates the soft attention to a frame $t$ while the softmax is applied over $\mbox{p}_{1..T}$. 
That is, the more similar the frame vector is to the other vectors, the higher its $\mbox{p}_t$ value is (\ie it is more likely to be the main scene of the video).
Finally, the frame encoding is simply obtained by a weighted sum of frame vectors by the attention values: $\mathbf g_\theta=\sum_{t=1}^T \mbox{p}_t \mbox{I}_t$.

\subsubsection{Adaptive Noise}
Each of 4,716 classes in the YouTube-8M dataset has a different number of video examples.
For example, the \textit{Car} class has about 800,000 examples, but the \textit{Air Gear} class has only 101 examples.
Let $n$ to be the number of labels associated with a video $\mathbf I$;
we introduce the adaptive noise structure to relieve the label imbalance problem as follows.
\begin{align}
& \mathbf{I}_t \leftarrow \mathbf{I}_t + \gamma \cdot \mbox{Z} \\ 
& \mbox{ where } \mbox{Z} \sim \mathcal{N}(0,I), \hspace{6pt} 
\gamma = \frac{1}{n} \cdot  \sum_{i=1}^n \frac{1}{S(y_i)}
\end{align}
\noindent where noise $\mbox{Z}$ is sampled from normal distribution, $y_i$ denotes the $i$-th label for video, and $S(y_i)$ is the number of video examples that have label $y_i$.
It means that we increase the generalization for small classes by adding more noise to their frame vectors.
We then make summation of the vectors of all frames as the output of $\mathbf g_\theta$, as done in the position encoding.


\subsection{Classification Layer}
\label{sec:classification_layer}
Classification Layer $h_\theta$ is defined as follows.
\begin{equation}
\label{eq:classification_layer}
\begin{cases}
h_\theta: \mathbb{R}^{T \times 1152} \to \mathbb{R}^{4,716}, &
\mbox{if } h_\theta \mbox{ is Many-to-Many} \\

h_\theta: \mathbb{R}^d \to \mathbb{R}^{4,716}, & \mbox{otherwise}
\end{cases}
\end{equation}
That is, by default, $h_\theta$ takes frame encoding $\mbox{O}_g$ as an input and outputs score $\mbox{O}_h$ for 4,716 classes.
Exceptionally, as shown in figure \ref{fig:overall}, the Many-to-Many model has $\mathbf{I}$ as its input, where $\mathbf{I}$ is untouched by video pooling layer $g_\theta$.
We have conducted experiments on the following three structures in this component. (see figure \ref{fig:class_layer}(a))
\subsubsection{Many-to-Many}
Unlike other models, Many-to-Many model has a frame vector $\mathbf{I}$ that is not touched by video pooling layer $g_\theta$.
It uses LSTM similar to that of \ref{sec:lstm}, but it calculates the score by attaching fully-connected layer to each output of each step in LSTM and average them, which is used as output $\mbox{O}_h$.
Since this model averages out the score in each frame, it has the temporal encoding ability of the RNN as well as the fact that the scores drawn in the more frequently appearing frames are reflected more. 
As a result, more effective video encoding could be performed.

\subsubsection{Variants of Mixture of Experts}
The Mixtures of Experts ~\cite{jordan1994hierarchical} model is a binary classifier that adaptively takes into account the scores of several experts corresponding to a class.
For one class, each expert $e_i$ has a probability value between 0 and 1, and gate $g_i$ represents the weight for each expert and is defined as follows.

\begin{align}
\label{eq:moe1}
e_i &= \sigma(\mathbf{w}_e^T \mbox{O}_g + \mathbf{b}_e) \\
\label{eq:moe2}
g_i &= \mbox{softmax}(\mathbf{w}_g^T \mbox{O}_g + \mathbf{b}_g)
\end{align}

where $\mathbf{w}_e$, $\mathbf{w}_g$ are $d$-dimensional vectors, scalars $\mathbf{b}_e, \mathbf{b}_g$ are biases, and softmax is performed for \{$g_1, ..., g_E$\}. ($E$ is number of experts.)
We extend this MoE model to multi-layer, and construct multiple fully-connected of defining probability and gate distribution of each expert.
For example, the gate distribution and the expert distribution of the 2-layer MoE model are defined as follows.

\begin{align}
\label{eq:moe3}
e_i &= \sigma(\mathbf{v}_{e}^T (\mathbf{W}_{e} \mbox{O}_g + 
\mathbf{b}^\prime_e) + \mathbf{b}_e) \\
\label{eq:moe4}
g_i &= \mbox{softmax}(\mathbf{v}_{g}^T (\mathbf{W}_g \mbox{O}_g + \mathbf{b}^\prime_g) + \mathbf{b}_g)
\end{align}

where each weight matrix $\mathbf{W}_{e}$,  $\mathbf{W}_g$ is denoted by $\mathbf{W}_e \in \mathbb{R}^{d^\prime \times d}$, $\mathbf{W}_g \in \mathbb{R}^{d^\prime \times d}$ and each $\mathbf{v}_{e}^T$, $\mathbf{v}_{g}^T$ means $d^\prime$-dimensional vector. $\mathbf{b}_e, \mathbf{b}_g, \mathbf{b}^\prime_e, \mathbf{b}^\prime_g$ are biases.
Finally, the score $\mbox{O}_h[i]$ for i-th class is determined by the weighted sum of each expert distribution; $\mbox{O}_h[i] = \sum_{i=1}^k e_i g_i$.

\subsubsection{Multi Layer Perceptron}
The Multi Layer Perceptron model is stack of Fully-Connected Layers, one of the most basic Neural Network structures.
We experimentally set the number of layers to 3 and apply Layer Normalization~\cite{ba2016layer}  to each layer.

\subsection{Label Processing Layer}
\label{sec:label_processing_layer}
Label Processing Layer $C_\theta$ is defined as follows.
\begin{equation}
C_\theta: \mathbb{R}^{4,716} \to \mathbb{R}^{4,716}
\end{equation}

This component is designed to reflect the correlation between the labels into the model.
For example, YouTube-8M, which includes \{\textit{Soccer, Football, Kick, Indoor soccer}\} and \{\textit{Super Mario Bros, Super Mario World, Super Mario bros 3, Mario Kart, Mario Kart 8}\}.
In order to take advantage of this property, we set up the label correlation matrix $\mbox{M}_c$ as follows by counting all the videos in the training set. (See Figure \ref{fig:class_layer}(b))
\begin{equation}
\mbox{M}_c \in \mathbb{R}^{4,716 \times 4,716} 
\end{equation}
where $\mbox{M}_c$ is correlation matrix  and the correlation value $\mbox{M}_c [i, j]$ between the i-th label and the j-th label is calculated higher as i-th label and j-th label appear together more in the same video.
Then, a new score is defined as follows to better reflect the correlation between labels through simple linear combination of matrix-vector multiplication.

\begin{equation}
\label{eq:label_correlation}
\mbox{O}_c = \alpha \cdot \mbox{O}_h + \beta \cdot \mbox{M}_c \mbox{O}_h  + \gamma \cdot \mbox{M}_c^\prime \mbox{O}_h
\end{equation}

Here,  $\mbox{M}_c$ is used as a fixed value, and $\mbox{M}_c^\prime$ is a trainable parameter initialized to the same value as  $\mbox{M}_c$. Scalar values $\alpha, \beta, \gamma$ are hyperparameters for model.

\subsection{Loss Function}
\label{sec:loss_function}

\subsubsection{Center Loss}
Center Loss was first proposed for face recognition task~\cite{wen2016discriminative} and expanded to other field because of its effectiveness for making dicriminative embedding feature~\cite{wang2017face}.
The purpose of the center loss is to minimize intra-class variations while maximizing inter-class variations using the joint supervision of cross-entropy loss and center loss.  
The original center loss was used for the single label classification problem and it is hard to exploit it in a multi-label classification problem. 

If we convert the multi label problem into a single label problem like ~\cite{de2009tutorial}, increment of centers according to the combination of labels is the simple expansion of center loss to a multi-label classification problem. 
However, this simple expansion is not suitable for YouTube-8M, because the number of combination for labels is too big to calculate. 
Therefore, we modified the center loss to suit the multi-label classification problem as follow.
\begin{align}
\label{eq:centerLoss}
\mathcal{L}_{c} &= \frac{1}{N} \sum_{i=1}^N \left \| \boldsymbol{e}_{i} - \boldsymbol{c}_{k} \right \|^2_{2},   k = {y}_{i} \\
\label{eq:totalLoss}
\mathcal{L}  &= \mathcal{L}_{s} + \lambda \mathcal{L}_{c}.
\end{align}
Where ${N}$ denotes the number of labels in one video. 
$\mathbf{e}_{i}$ denotes a embedding vector from penultimate layer, 
$\mathbf{c}_{k}$ denotes the ${y}_{i}$-th corresponding class center for each $\mathbf{e}_{i}$.
The $\mathcal{L}_{s}$ is the cross-entropy loss, $\mathcal{L}_{c}$ is the center loss.
A scalar $\lambda$ is a hyperparameter for balancing the two loss functions.

\subsubsection{Pseudo-Huber Loss}
The Huber Loss is a combination of L2 Loss and L1 Loss, which allows the model to be trained more robustly on noise instances.
In the case of the YouTube-8M, there are classes with very few instances due to the label imbalance problem, and Huber Loss is designed to better learn instances belonging to these classes.
For a simple, differentiable form, we use the Pseudo-Huber Loss function, a smooth approximation of the Huber Loss, as follows.
\begin{equation}
\mathcal{L}_c = \delta^2 (\sqrt{1 + (\mathcal{L}_{CE}/\delta)^2}-1)
\end{equation}
where $\mathcal{L}_{CE}$ means that Cross-Entropy Loss between our prediction $\mbox{O}_c$ and ground-truth label $y$, $\delta$ means hyperparameter for model.

\subsection{Training}
\label{sec:training}

For training our model, we choose the Adam~\cite{kingma2014adam} optimizer using a batch size of 128, with learning rate = 0.0006, $\beta_1 = 0.9$, $\beta_2 = 0.999$, $\epsilon = 1e-8$.
We also apply learning rate decay with the rate = $0.95$ for every 1.5M iterations.
We train the model for 5 epochs with no early stopping. 
We use both official training and validation data that YouTube-8M public dataset provides for training our models.

\section{Experiments}
\label{sec:experiments}

We use the test data from the Kaggle competition: Google Cloud \& YouTube-8M Video Understanding Challenge to measure the performance of the model.
The source for our model is publicly available\footnote{https://github.com/seilna/youtube-8m}.

\subsection{Experimental Setting}
One of the greatest features of our model is that we have three issues to solve the YouTube-8M classification task; i)temporal encoding for video, ii)label imbalance problem, iii)correlation between labels, and we have tried several variations on each component by dividing the model pipeline into four components; i)Frame Encoding, ii)Classification Layer, iii)Label Processing Layer, iv)Loss Function.
In addition, except in the Many-to-Many model in our pipeline, each component is completely independent of one another, so it is a very good structure for experimenting with a number of variant model combinations.
However, it is impossible to do brute-force experiments because the number of trials is too great to test all combinations of variations that each component can try.
Thus, we took the greedy approach, fixed the remaining 3 components in each component, and experimented the several  structures only for that component, choosing the best-performing structure for each component.
We used Google Average Precision\footnote{https://www.kaggle.com/c/youtube8m\#evaluation} (GAP)@20 as a metric to measure the performance of the model.

\begin{table}[]
\centering
\begin{tabular}{lclll}
\hline
Method           & \multicolumn{4}{c}{GAP@20} \\ \hline
\verb'LSTM'               & \multicolumn{4}{c}{0.811}  \\
\verb'LSTM-M'              & \multicolumn{4}{c}{0.815}  \\
\verb'LSTM-M-O'            & \multicolumn{4}{c}{\textbf{0.820}}  \\
\verb'LSTM-M-O-LN'         & \multicolumn{4}{c}{0.815}  \\ \hline
\verb'CNN-64'              & \multicolumn{4}{c}{0.704}  \\
\verb'CNN-256'             & \multicolumn{4}{c}{0.753}  \\
\verb'CNN-1024'            & \multicolumn{4}{c}{-}      \\ \hline
\verb'Position Encoding'   & \multicolumn{4}{c}{0.782}  \\ 
\verb'Indirect Clustering' & \multicolumn{4}{c}{0.801}  \\ 
\verb'Adaptive Noise' & \multicolumn{4}{c}{0.782}  \\
\hline
\end{tabular}
\vspace{6pt}
\caption{The results on various transformation structures in Video Pooling Layer $g_\theta$ component.
}
\label{tab:component1}
\end{table}

\subsection{Quantitative Results}
\label{sec:quatitative}
\subsubsection{Video Pooling Layer}
There are five transform structures in Video Pooling Layer $g_\theta$; i) variants of LSTM, ii) CNN, iii) Position Encoding, iv) Indirect Clustering, v) Adaptive Noise.
To select the most suitable structure for  $g_\theta$, we fixed the rest of the component $h_\theta$ to the MoE-2 model, not $C_\theta$ used, and $\mathcal{L}_c$ used the cross entropy loss.
Based on these settings, the results for each of the five structures are shown in table \ref{tab:component1}.

\verb'LSTM' simply uses the last hidden state, \verb'LSTM-M' concatenates $\mbox{M}_l$, \verb'LSTM-M-O' concatenates both $\mbox{M}_l$ and $\mbox{O}_l$, and \verb'LSTM-M-O-LN' is a model that applies Layer Normalization for each layer of LSTM.
\verb'CNN-64', \verb'CNN-256', and \verb'CNN-1024' refer to models with 64, 256, and 1024 as the output channels of CNN, respectively.

The results show that the LSTM method has the best performance for encoding frames.
Within LSTM, the higher the utilization of the internal information, the higher the performance, indicating that higher performance can be expected if more information can be extracted from other methods such as skip connection in LSTM.
On the other hand, unlike the expectation, the LSTM with Layer Normalization has a lower performance than that which is not.
Of course, Layer Normalization had the advantage of stable and fast convergence even when 20-30 times learning rate was applied, but when compared to the final performance alone, performance was not good.

CNN showed a very poor performance unexpectedly.
As CNN's output channel increased, performance was on the rise, but channels larger than 256 were not experimented because of the memory limitations of the GPU.
While we can expect that CNN will perform well for larger channels, another problem with CNN is the dramatic increase in computation cost as channels increase.
However, if CNNs with different hyperparameters record similar performance to LSTM, CNN is likely to be used as good pooling method because it has the advantage that the convolution operation is fully parallelizable.

The Position Encoding model showed lower performance than LSTM, and one of the possible reasons is that the sequence modeling power of the model is weaker than LSTM.

The Indirect Clustering model has lower performance than LSTM, but it performs better than the Position Encoding model.
This suggests that the assumptions we have made (the importance of the main scene in video classification) are not entirely wrong, suggesting that we need a model that can cover this issue more delicately.
Also, temporal encoding of LSTM is as important as considering main scene in video classification.

Adaptive Noise do not lead to a significant performance improvement, indicating that a more sophisticated approach to the label imbalance problem is required.

\begin{table}[]
\centering
\label{tab:component2_g}
\begin{tabular}{lcl}
\hline
Method              & \multicolumn{2}{c}{GAP@20}         \\ \hline
\verb'Many-to-Many'        & \multicolumn{2}{c}{0.791}          \\ \hline
\verb'2 Layer MoE-2'       & \multicolumn{2}{c}{0.424}          \\
\verb'2 Layer MoE-16'      & \multicolumn{2}{c}{0.421}          \\ \hline
\verb'3 Layer MLP-4096'    & \multicolumn{2}{c}{0.802}          \\
\verb'3 Layer MLP-4096-LN' & \multicolumn{2}{c}{\textbf{0.809}} \\
\hline
\end{tabular}
\vspace{6pt}
\caption{The results on various transformation structures in Classification Layer $h_\theta$ component.}
\end{table}

\subsubsection{Classification Layer}
The Classification Layer $h_ \theta$ has three variants: i)Many-to-Many, ii)Multi-Layer MoE, iii)MLP.
To select the best method for $h_\theta$, we fix the remaining component $g_\theta$ with Indirect Clustering, $C_\theta$ is not used, and $L_c$ uses cross entropy loss. (However, in the many-to-many model, $g_\theta$ is not used according to the definition.)
Based on these settings, the results for each of the three structures are shown in Table 3.

\verb'2 Layer MoE-2' refers to a 2 layer MoE model with 2 experts, and \verb'2 Layer MoE-16' refers to a 2 layer MoE model with 16 experts.
\verb'3 layer MLP-4096' is MLP structure with 4096 dimension of each hidden layer and 3 hidden layers, and \verb'3 layer MLP-4096-LN' applies layer normalization to each layer.

In constructing the classification layer, the MLP structure showed the best performance among the three methods.
The Many-to-Many model is an LSTM-based model, but has a lower performance than the basic LSTM structure in Table 2, indicating that the Many-to-Many framework in video classification is not always a good choice.

On the other hand, disappointingly, the Multi Layer MoE model showed severe overfitting in only two layers.
As expected, this requires a lot of parameters to create an intermediate level of embedding, which is also needed for each experts, resulting in overfitting for many parameters.
We experimented with different number of layers and hidden layer dimensions for the MLP structure, but 4096 dimensions of 3 layers showed the highest performance.
Unusual is that, when layer normalization was applied, the MLP model showed improved performance unlike LSTM, and did not require more delicate hyperparameter tuning.

\subsubsection{Label Processing Layer}

\begin{table}[t]
\centering
\begin{tabular}{lclc}
\hline
\multicolumn{1}{l}{Method} & \multicolumn{2}{c}{GAP@20}         \\ \hline
\verb'MoE - (1.0, 0.3, 0.0)'      & \multicolumn{2}{c}{0.784}          \\
\verb'MoE - (1.0, 0.1, 0.0)'      & \multicolumn{2}{c}{0.787}          \\
\verb'MoE - (1.0, 0.0, 0.1)'      & \multicolumn{2}{c}{0.788}          \\
\verb'MoE - (1.0, 0.01, 0.0)'     & \multicolumn{2}{c}{\textbf{0.790}}          \\
\verb'MoE - (1.0, 0.0, 0.01)'    & \multicolumn{2}{c}{\textbf{0.790}}          \\
\verb'MoE - (1.0, 0.01, 0.01)'    & \multicolumn{2}{c}{0.788} \\
\hline
\end{tabular}
\label{tab:component3}
\vspace{6pt}
\caption{The results on various hyperparameters in Label Processing Layer $C_\theta$ component.}
\end{table}

To select the best method for $C_\theta$, we fix the remaining component $g_\theta$ with Indirect Clustering, $h_\theta$ using the MoE-16 model and $\mathcal{L}_c$ using the cross entropy loss.
As defined in Equation \ref{eq:label_correlation}, the Label Processing Layer $C_\theta$ uses pre-computed correlation matrix $\mbox{M}_c$.
$\alpha, \beta$, and $\gamma$ values ​​can be used to control the degree to which the correlation matrix affects the class score. 
An experiment is shown in Table 4.

As a result, performance dropped for all $\beta$ and $\gamma$  values ​​greater than zero.
In other words, when using the label correlation matrix for the purpose of score update, it showed rather low classification performance.

The possible reason for the result may be that our model is too naive to reflect the correlation label prior and that label correlation in the YouTube-8M Dataset is not strong enough to improve the performance of the classification task.

\subsubsection{Loss Function}
\begin{table}[]
\centering
\label{my-label}
\begin{tabular}{lcl}
\hline
\multicolumn{1}{l}{Method} & \multicolumn{2}{c}{GAP@20}         \\ \hline
$\mathcal{L}_{CE}$                      & \multicolumn{2}{c}{0.798}          \\
$\mathcal{L}_{CE}$ + $\mathcal{L}_c$($\lambda=0.001$)                & \multicolumn{2}{c}{0.799}          \\ \hline
$\mbox{Huber}_{CE}$($\delta=0.5$)                  & \multicolumn{2}{c}{\textbf{0.803}} \\
$\mbox{Huber}_{CE}$($\delta=1.0$)                   & \multicolumn{2}{c}{0.801}          \\
$\mbox{Huber}_{CE}$($\delta=2.0$)                   & \multicolumn{2}{c}{0.798}          \\
$\mbox{Huber}_{CE}$($\delta=3.0$)                   & \multicolumn{2}{c}{0.794} \\
\hline
\end{tabular}
\vspace{6pt}
\caption{The results on variants in loss function $L_c$ component.}
\end{table}

To select the most suitable loss for $\mathcal{L}_c$, we fix the remaining component $g_\theta$ with Indirect Clustering, $h_\theta$ uses the MoE-16 model, and $C_\theta$ is not used.

Table 5 shows the performance depending on whether or not the center loss and the pseudo-huber loss are used.
$\mathcal{L}_{CE}$ is the cross entropy loss, $\mathcal{L}_c$ is the center loss, and $Huber_{CE}$ is the Pseudo-Huber loss for cross entropy term.

The results show that, firstly, using the center loss term $\mathcal{L}_c$ gives little performance gain.
Considering that we apply the center loss term to the multi-label, the performance should be improved if the labels are correlated with each other.
Here we get another insight for the correlation label prior, which is that the correlation does not exist as strongly as to achieve performance improvement.
Second, Huber Loss proved its ability to cover noise data somewhat from the YouTube video annotation system\footnote{https://www.youtube.com/watch?v=wf\_77z1H-vQ} by recording a relatively clear performance improvement.

\subsection{Ensemble Model}
Based on the experiments we conducted in section \ref{sec:quatitative}, we recorded test performance of 0.820 as a single model combining LSTM-M-O, MoE-2, and HuberLoss.(Due to GPU memory limitation, we could not apply MoE-16 or Multi Layer MLP model to LSTM based model)
In addition, our ensemble model, which is a simple average of the scores of several models, showed a test performance of \textbf{0.839} and ranked 8th in the kaggle challenge.
The interesting thing is that when we assemble several models, we did not get a significant performance increase if we combine the better single models.
Rather, for increasing ensemble model's performance, the models incorporated should be as diverse as possible.

\section{Conclusion}
\label{sec:conclusion}
We defined three issues that need to be covered in order to solve the YouTube-8M Video Classification task, and we divided the model pipeline into four components and experimented with the various structures to solve issues in each component.
As a result, almost all of the deformed structures tried to perform better than the baseline, and their ensemble model recorded \textbf{0.839} in the test performance and 8th in the kaggle challenge.
We also provided insights on the structures we tried on each component, on what roles each structure plays, and why they work well or poor.
Based on this insight, we will explore ways of better frame encoding or use more elegant label correlation priorities as our future work.

{\small
\bibliographystyle{ieee}
\bibliography{egbib}
}

\end{document}